\documentclass{article}
\usepackage{ijcai24}

\usepackage{times}
\usepackage{soul}
\usepackage{url}
\usepackage[hidelinks]{hyperref}
\usepackage[utf8]{inputenc}
\usepackage[small]{caption}
\usepackage{graphicx}
\usepackage{amsmath,amssymb,amsfonts}
\usepackage{booktabs}
\usepackage{algorithm}
\usepackage{algorithmic}
\usepackage{multirow}
\usepackage[switch]{lineno}

\usepackage{natbib}
\usepackage{enumitem}
\usepackage{xcolor}

\usepackage{tikz}
\usepackage[edges]{forest}
\definecolor{hiddendraw}{RGB}{205, 44, 36}
\definecolor{hidden-blue}{RGB}{194,232,247}
\definecolor{hidden-orange}{RGB}{243,202,120}
\definecolor{hidden-yellow}{RGB}{242,244,193}

\usepackage{makecell}
\usepackage{colortbl} 
\definecolor{LightGreen}{rgb}{0.8, 1, 0.8} 

\title{Foundation Model Sherpas: 
\\
Guiding Foundation Models through Knowledge and Reasoning}

\author{
Debarun Bhattacharjya
\and
Junkyu Lee
\and
Don Joven Agravante
\and
Balaji Ganesan
\And
Radu Marinescu
\affiliations
IBM Research\\
\emails
debarunb@us.ibm.com, \{Junkyu.Lee, Don.Joven.R.Agravante\}@ibm.com, \\ bganesa1@in.ibm.com, radu.marinescu@ie.ibm.com
}
\begin{document}

\maketitle

\begin{abstract}
Foundation models (FMs) such as large language models have revolutionized the field of AI by showing remarkable performance in various tasks. However, they exhibit numerous limitations that prevent their broader adoption in many real-world systems, which often require a higher bar for trustworthiness and usability. Since FMs are trained using loss functions aimed at reconstructing the training corpus in a self-supervised manner, there is no guarantee that the model's output aligns with users' preferences for a specific task at hand. In this survey paper, we propose a conceptual framework that encapsulates different modes by which agents could interact with FMs and guide them suitably for a set of tasks, particularly through knowledge augmentation and reasoning. Our framework elucidates agent role categories such as updating the underlying FM, assisting with prompting the FM, and evaluating the FM output. 
We also categorize several state-of-the-art approaches into agent interaction protocols, highlighting the nature and extent of involvement of the various agent roles. The proposed framework provides guidance for future directions to further realize the power of FMs in practical AI systems. 
\end{abstract}

\section{Introduction}

Foundation Models (FMs) and in particular Large Language Models (LLMs) have revolutionized AI~\citep{bommasani,han2021pretrained}.
They are currently at the core of several products that touch millions of users on an almost daily basis. However, in spite of all their promise, FMs/LLMs are known to suffer from important limitations that prevent their broader adoption, including lack of traceability, interpretability, formal reasoning,  ability to easily incorporate  domain knowledge, and generalizability to new environments.

To alleviate some of the well-known issues with FMs, there has been a trend towards trying to ensure \emph{alignment} of FMs -- a term borrowed from the AI safety literature~\citep{amodei2016concrete}. Since FMs are trained through loss functions that are intended to generate the training corpus in a self-supervised manner, there is no guarantee that the model’s output aligns with users' preferences for the output, in terms of the task at hand. A recent poster child for this general direction is the reinforcement learning with human feedback (RLHF) paradigm of ChatGPT, where human feedback is first used to learn reward models of the generated output and subsequently leveraged for generating future output~\citep{stiennon}.

There are however numerous challenges that must be addressed
for FMs to be effectively aligned in many practical systems.
Firstly, there may be multiple users and user personas that interact
with such systems, and each user (or user type) may have different
objectives and preferences. Furthermore, datasets that are leveraged
for pre-training may be outdated or generally problematic
without additional guardrails due to concerns such as around bias
and privacy. This is important in business problems or domains
such as healthcare where incorporating domain-specific knowledge
in various forms is essential. Another issue is that of robustness;
practical systems may need to deal with shifting data distributions,
and perhaps changing external factors such as regulations. FMs
such as LLMs are essentially token generators, and changing the
nature of generation by trial-and-error approaches through prompt
engineering or updating the FM through simple human feedback
alone is unlikely to yield long-lasting performance and build trust.

In this survey paper, 
we introduce a conceptual framework where agents/modules that interact with FMs behave as \emph{sherpas}, in the sense of guiding FMs to tackle user-specified tasks. 
We emphasize how such agents can leverage \emph{knowledge} and \emph{reasoning} in particular. 
While there are several recent survey papers on related subjects, such as prompt engineering in FMs~\citep{liu2021pretrain,dong2023survey}, knowledge enablement in FMs~\citep{wei2021knowledge,zhen2022,pan2023unifying}, reasoning in FMs~\citep{huang2023,qiao2023reasoning}, and instances of augmentation of FMs~\citep{mialon2023augmented}, our 
framework differs in that it emphasizes the \emph{roles} played by various agents as they interact with FMs to
pursue user-specified tasks. Emphasizing an agent’s role helps in
defining its motivating objectives and goals.

In the remainder of this
paper:
\begin{itemize}[noitemsep,nolistsep,leftmargin=*]
\item We distinguish the scope of our ideas from some ongoing efforts that incorporate
agents into systems with FMs.
\item We introduce the sherpas framework, which categorizes the purpose
of various agents, going well beyond the generic role of
‘augmentation’. We describe how the
specific roles in the framework connect to selected recent work; this distinguishes our survey from other related surveys, which typically focus solely on either a specific role (such as involving prompting), or technical approach for augmentation (such as the deployment of knowledge graphs). 
\item 
We show how selected state-of-the-art  approaches invoke a combination of some of the agent roles from the framework, through a categorization based on the extent of formal reasoning involved. We also highlight further opportunities for new types of interaction and collaboration between agents,
with or without human involvement.
\item We briefly motivate a vision for a future where the proposed guiding
agents are truly autonomous agents rather than system modules,
in that they exhibit key behavior such as being proactive, reactive,
interactive, and collaborative.
\end{itemize}

While our exposition hinges primarily around
LLMs, the ideas apply generally to all FMs. This includes models
trained on other data modalities beyond text, such as images, video,
audio, code, tabular data, knowledge graphs, time series data, spatiotemporal
data, and even multi-modal data.




\section{Agents and Foundation Models}


There is an increasing interest in work at the intersection of agent-based perspectives and systems incorporating FMs. For instance, an approach inspired by Minsky's `society of minds' 
leverages multiple language model instances (or agents) that individually propose and jointly debate their responses and reasoning processes to arrive at an answer~\citep{du2023improving}. 
Another approach pursues a communicative agent framework called ``role playing'' that utilizes inception prompting to guide chat agents in task completion, generating conversational data for studying multi-agent cooperative behaviors~\citep{li2023camel}. 

Recent strides have also been made in \emph{orchestration} frameworks, which seamlessly integrate FMs into various applications and workflows. An example is Langchain~\citep{langchain}, which enables systems that couple LLMs with tools and APIs such as code execution and knowledge retrieval. 
Another recent software framework effort that complements our own vision is AutoGen~\citep{wu2023autogen} -- a programmatic orchestration layer that allows developers to build FM-based applications using a collection of `agents' 
to solve tasks more effectively. These agents can operate in various modes and involve combinations of several different LLMs, human input, as well as employ tools and APIs (similar to Langchain). 
Our perspective here is consistent with these frameworks but also emphasizes agent roles beyond that of orchestration. Furthermore, we stress the autonomy of agents and what we consider to be desirable agent-like behavior; we do not regard tools/APIs to be agents in that sense, unlike these other frameworks.

\section{Agent Roles in the Sherpas Framework}

\begin{figure*}
\centering
{\includegraphics[width=0.7\textwidth]{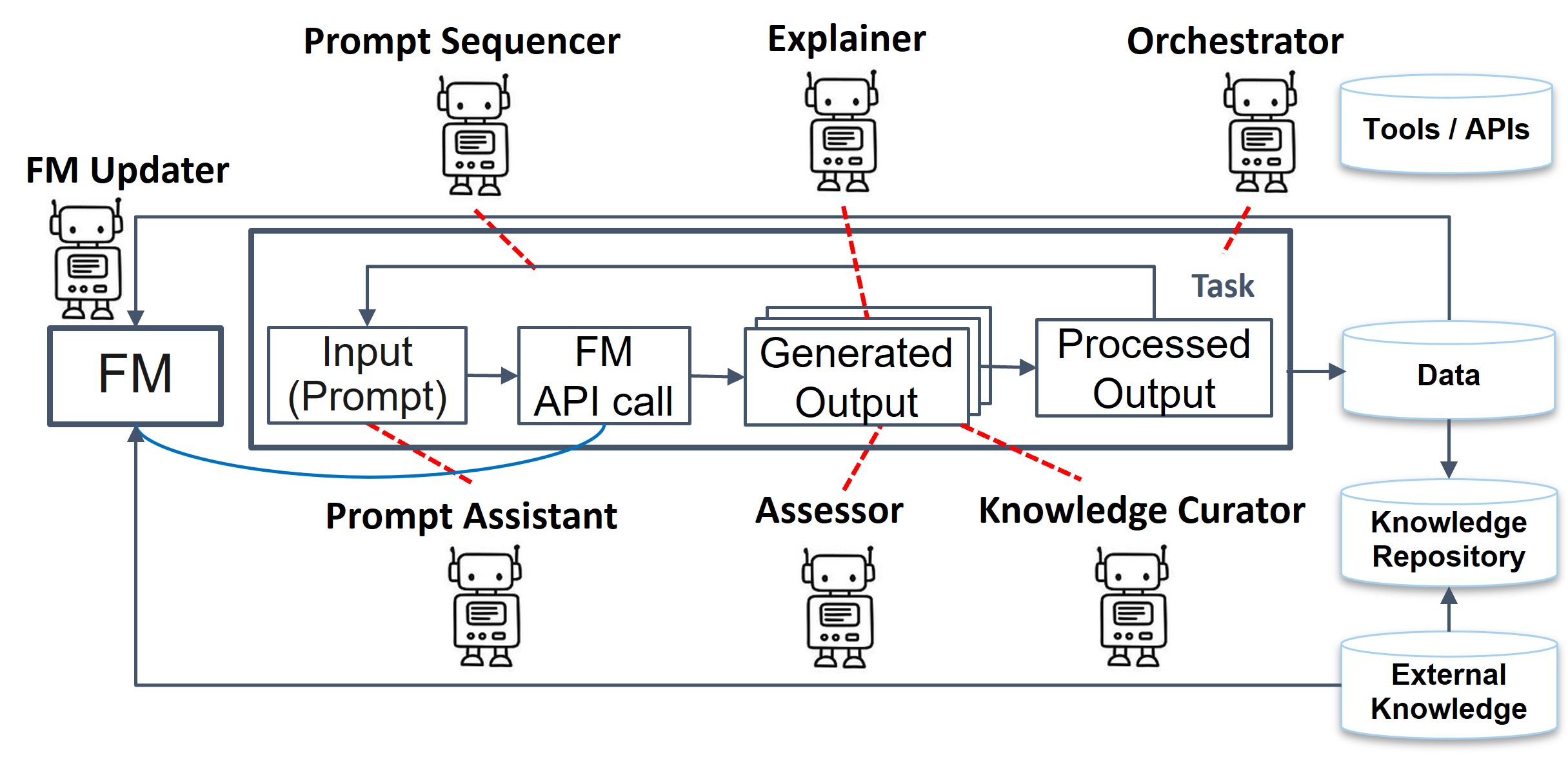} }
\caption{The sherpas framework for guiding FMs, showing various agent categories with respect to their typical points of interaction with the FM as it executes or assists completion of a set of tasks.}
\label{fig:framework}
\end{figure*}

Figure~\ref{fig:framework} depicts components of the  sherpas framework, distinguishing between various types of agents in how they guide an FM as it addresses a set of tasks. 
In general, a task is executed as follows: an input (prompt) such as a sequence of words is provided to an FM, which is invoked to generate an output (completion), which may be subsequently evaluated or otherwise processed. 
Solving a task may require multiple FM calls, and there may be an opportunity to leverage external knowledge and/or reasoning via one or more sherpas.
In the remainder of this section, we categorize sherpa roles in the task-solving process and provide some pointers to related work. 
Sherpas may be humans or machines or involve some combination thereof.

\begin{figure*}
\centering
{\includegraphics[width=0.85\textwidth]{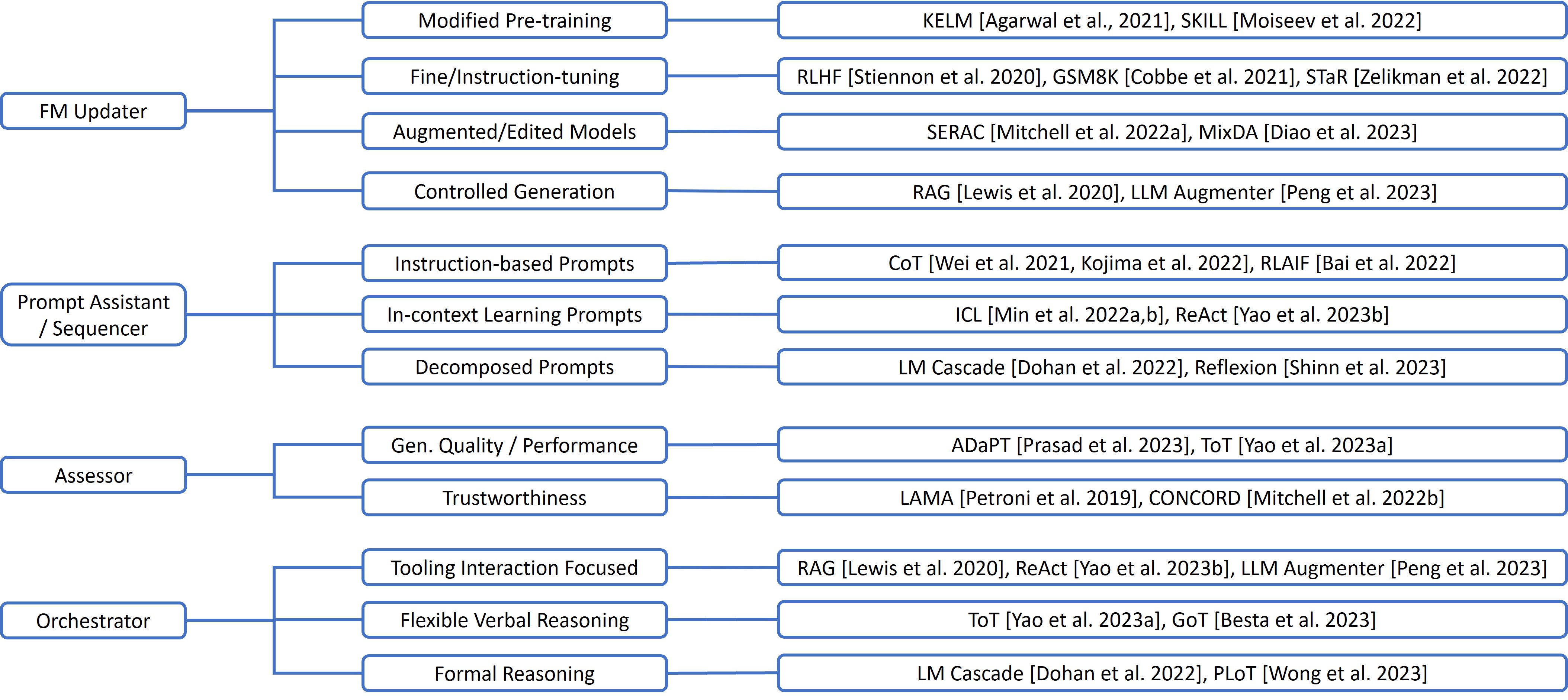} }
\caption{A relatively high-level taxonomy of four major sherpa categories, along with some illustrative example references.}
\label{fig:taxonomy}
\end{figure*}

In the remainder of this section, we describe the purpose of each sherpa category and provide a coarse sub-categorization; these are intended to be sufficiently illustrative but not fully comprehensive, given the space limitations and rapidly expanding literature on foundation models. 
Note that in most current work, many of these functions are performed by modules of systems.
Figure~\ref{fig:taxonomy} provides a summary depiction of the high-level taxonomy, along with some example references.

\subsection{FM Updaters}


\texttt{FM Updaters} are sherpas that modify the underlying transformer or how the FM generates tokens by using external memory or models, 
to improve performance or better alignment with user preferences for a specific task. 
We provide a few illustrative examples from the following four broad categories. 



\subsubsection*{Modified Pre-training}

Various approaches modify the manner in which an FM is pre-trained.
An important family of these agents performs \emph{knowledge infusion} into an FM, which has been shown to help in downstream tasks. 
For instance, additional pre-training of a flan-T5 model on sentences generated from Wikidata triples~\citep{agarwal2021kelm} or using the triples directly~\citep{moiseev2022skill} helped with question-answering. 

\subsubsection*{Fine/Instruction-tuning}

Perhaps the most common approach is through \emph{fine-tuning} a pre-trained FM with labeled data. 
An important sub-class of such agents performs fine-tuning using data that is generated from intelligent prompt choices. For instance, a series of work uses intermediate verbal reasoning (`rationales') in prompts to improve task performance~\citep{rajani-etal-2019-explain,cobbe2021training,zelikman2022star}. 
An approach popularized by ChatGPT leverages reinforcement learning to learn a policy network for output generation with reward signals collected from feedbacks ~\citep{stiennon,bai2022constitutional,ge2023openagi}.
We note that such approaches also involve \texttt{Prompt Assistant} sherpas.

\subsubsection*{Augmented/Edited Models}

A broad category of agents modifies generation through architectural augmentation or proxies during inference, or by directly editing the model weights. For instance, adapters have been proposed to achieve similar effects as compared to knowledge infusion, using parameter efficient fine-tuning rather than additional pre-training~\citep{wang2021k,diao2023mixtureofdomainadapters}. 
An alternate approach is \emph{knowledge editing}, which involves modifying the weights of the original model to incorporate any changed facts~\citep{de2021editing,mitchell2022memory}. 
%
Proxy tuning~\citep{liu2024tuning} is an approach that emulates a black-box LLM without changing the model's weights.


\subsubsection*{Controlled Generation}

Some agents perform \emph{controlled generation} by modifying the manner in which tokens are generated without necessarily changing the underlying FM. 
For instance,
\citet{zhong2023mquake} propose an alternate approach which does not update the LLM but instead 
generates a tentative response 
and checks if the output contradicts updated facts in external memory. 
%
%


\subsection{Prompt Assistants and Sequencers}

Prompting is the primary mode of interacting with FMs, and \texttt{Prompt Assistants} help with the process of \emph{prompt engineering}.  
We refer to these agents as \texttt{Prompt Sequencers} when multiple calls are made to an FM for a task, requiring several prompts to be composed and sequenced.
Empirical studies have shown that good prompts offer FMs highly informative content and specific semantic meanings in pre-defined templates.
Two broad categories of basic prompting have emerged: instruction-based and example-based. We also highlight decomposed prompts as an important evolution of instruction-based prompts.

\subsubsection{Instruction-based Prompts}

These prompts contain explicit directions for what the user feels the FM should do. This is very intuitive and the main mode of using FMs. It is perhaps also the main reason for the widespread use of LLMs, because of how easy it is to write instructions in natural language.
Prompt instructions can be further subdivided into two sub-classes: task-prompts and system-prompts. Task-prompts are those which contain direct instructions about a task to accomplish (e.g. ``write a poem''). On the other hand, system-prompts offer broader guidelines or style (e.g. ``pretend you are a fantasy novel writer''). Constitutional principles in~\citet{bai2022constitutional} are another example of system prompts.
\texttt{Prompt Assistants} often use both of these sub-classes together for better effect. 

\subsubsection{In-context Learning Prompts}
The second major category of prompting uses examples of the task in the prompt. Instead of explicit instructions, these examples contain a pattern of the implicit instruction that FMs must follow. This is often referred to as \emph{in-context learning} and it has spurred a lot of research~\citep{min;acl;2022,min;emnlp;2022,rubin-etal-2022-learning}.
In tasks where a small dataset or some labeled examples are available, these can be directly used in the prompt as a source of knowledge. 
Chain-of-thought (CoT) prompting~\citep{wei2021knowledge,kojima;neurips;2022} is an important part of in-context learning prompts, where the research focus has been on elicitive prompts that guide FMs by decomposing a task into smaller tasks. 

\subsubsection{Decomposed Prompts}

Extending the CoT idea, \texttt{Prompt Sequencers} are frameworks that sequentially call FMs to solve sub-tasks. The decomposition itself can also be done by FMs~\citep{zhou2023least, khot2022decomposed}. 
Taking this idea even further, the sub-task decomposition can be turned into graphical models, linking this line of research to formal reasoning methods. For example, language model cascades~\citep{dohan2022language} proposes a probabilistic programming language paradigm that treats question-answering tasks as a graphical model over string-valued random variables for prompt inputs and LLM outputs, thereby unifying various prompt-based reasoning approaches~\citep{wei2023chainofthought,creswell2022selection}.

\texttt{Prompt Assistants} typically combine  different strategies because empirically they often synergize. In the broader picture, prompts are also the primary way to connect with other agents; this features more heavily in Section~\ref{sec:Agent Interaction Protocols}.

\subsection{Assessors and Explainers}


\texttt{Assessors} evaluate output generated from an FM 
along one or more \emph{attributes} for a task, including but not limited to measures of quality of generation, such as fluency, or those for trustworthiness, such as harmfulness, factuality,  etc. Such attributes should ideally have the following properties: 1) they should be pertinent to the users' values for the task(s), 2) they should be clearly defined, and 3) they should be measurable. 
Human feedback is of course a valuable source for evaluating such attributes, but there is also a recent trend of leveraging other models (sometimes other FMs) as assessors.
We distinguish between selected prior work based on the attribute being assessed. 

\subsubsection*{Generation  Quality / Performance Related}

There are several well-known  attributes and associated metrics that are intended to assess various facets of the quality of an FM's generated output. For LLMs, these include attributes of the generated language, such as \emph{fluency} (as measured by perplexity), \emph{coherence} (as measured by BLEU/ROUGE), \emph{relevance} or \emph{context awareness} (as measured by the BERT score or mutual information), or the \emph{ambiguity} of the output (as measured by accuracy). 
Some assessors try to evaluate metrics that are pertinent to the performance of the task at hand. For instance, they may assess whether or not FM generated code will \emph{execute or fail}~\citep{prasad2023adapt}, or measure the \emph{progress} made by the current state from a sequence of generated FM output with regard to whether it has the potential to answer the user's query~\citep{yao2023tree}.

\subsubsection*{Trustworthiness Related}

The following attributes are pertinent to help gauge the trustworthiness of the generated output:

\paragraph{Harmfulness/Toxicity.} 

FMs such as LLMs are known to occasionally generate content that may be deemed toxic or harmful by users. 
The RLAIF approach~\citep{bai2022constitutional} extends reinforcement learning from human feedback (RLHF) by generating harmful datasets through chain-of-thought prompting using a pre-defined collection of critique rules; the intention is to generate output that is deemed to be helpful yet harmless. In this work, the assessed score is based on crowd worker preference.

\paragraph{Factuality/Faithfulness.}

Generating output that is factual or faithful relative to some knowledge reference is often beneficial in practice.
For large-scale repetitive evaluation of FM outputs,
leveraging external knowledge such as knowledge graphs can be fruitful, since they offer easily interpretable and structured information. 
LAMA~\citep{petroni2019language}
presents a method for probing LLMs with knowledge graphs for measuring and improving the factuality of LLMs.
%

\paragraph{Consistency.}
ConCORD~\citep{mitchell2022enhancing} is an example of an approach that evaluates the consistency of FM generated output. 
It formulates a self-consistency measure as a compatibility score in a probabilistic model over the candidate question-answer pairs queried from LLMs. 
Note that consistency and factuality are closely related, as consistency asks for invariant outputs relative to semantically similar rephrased inputs,
while factuality requires consistent behavior relative to ground truth outputs~\citep{elazar2021}.

\paragraph{Uncertainty/Confidence.}
An important class of \texttt{Assessors} performs \emph{uncertainty quantification} to estimate measures such as the variability or confidence of the FM output~\citep{lin2022teaching,kuhn2022semantic}.
These have been further categorized as \emph{verbalized}, where the FM provides confidence scores directly, or \emph{non-verbalized}, where confidence is inferred by other means such as gauging the consistency of the output after paraphrasing the input prompt.

\paragraph{Hallucination.}
FMs are widely known to generate output that are considered hallucinations (or confabulations).
\citet{huang2023survey} categorizes them as either
input-conflicting, where the output deviates from user-provided source
input, context-conflicting, where the output conflicts with previously
generated information by itself, and fact-conflicting, where the output is not faithful to established world knowledge. Note that fact-conflicting hallucination can be assessed by factuality measures, as described previously.

\paragraph{Bias.}
Since an FM's training data often reflects stereotypes and discriminatory beliefs about gender, race, and other societal factors, assessing such biases could be useful in deployments such as chat bots~\citep{zhao2023gptbias}.

\texttt{Explainers} are an important class of agents related to \texttt{Assessors}, and help provide explanations about the output of FMs; we refer the reader to a complementary survey paper which reviews relevant advances~\citep{zhao2023explainability}.
Explainability has another connection to assessment of FM output -- critical applications sometimes need an explanation for the evaluation, not just the evaluation itself. For instance, a user may appreciate receiving records in a knowledge base that provide evidence for a factuality assessment.

\subsection{Knowledge Curators}

Many AI systems that incorporate FMs require access to sources of knowledge, such as structured information in a relational database or an unstructured corpus of domain knowledge. This necessitates the involvement of an often neglected category of agents, which curate knowledge from appropriate
sources with the intent of supporting the FM towards handling the task(s) of interest.

\texttt{Knowledge Curators} play a particularly important role through their interaction with other guiding agents. For instance, they are needed for \texttt{Assessors} that rely on external knowledge to provide a reference for factuality~\citep{petroni2019language}. They are also crucial for many \texttt{FM Updaters}.
For instance, they are needed by retrieval-augmented generative models~\citep{lewis_RAG}, which combine pre-trained language models with a non-parametric memory (such as a dense vector index of Wikipedia). 


We also include in this broader category agents that are able to suitably extract information from FMs, whenever it is beneficial for the application. 
Since FMs memorize a lot of facts, this class of \texttt{Knowledge Curators} helps acquire and maintain an FM's inherently stored knowledge in more structured forms. Due to the prevalence and simplicity of knowledge graphs, 
there have been numerous efforts to extract them from LLMs~\citep{bosselut-etal-2019-comet,west_distillation,hao2023bertnet}. There is also upcoming research on acquiring other representations of knowledge, such as finite-state automata~\citep{yang2023automatonbased} and causal models~\citep{arsenyan2023large,kıcıman2023causal,long2023large,shi2023language}. 
\texttt{Knowledge Curators} of this type can aid an AI system by being suitably leveraged in future relevant tasks.

\subsection{Orchestrators}

Real-world AI systems often rely on \texttt{Orchestrators} (or Controllers) to manage potentially complex workflows. Current approaches in the literature highlight interactions with external data sources, tools and applications in particular.
We have previously described the burgeoning interest in frameworks to manage such workflows, including Langchain and AutoGen.
We classify these into three high-level categories, primarily based on the manner in which formal reasoning is mimicked or leveraged for the process of orchestration.



\subsubsection*{Tooling Interaction Focused}

An important line of research studies the use of LLMs to directly generate domain-specific actions or plans in interactive environments. Many of these efforts attempt to combine `verbal reasoning' and `acting' for solving diverse language reasoning and decision making tasks; since they put more of the onus on the LLM for managing work flows, this reduces the load of the \texttt{Orchestrator}, which is typically only left with 
executing actions such as calling tools/APIs. For instance, the ReAct framework \citep{yao2023react} prompts an LLM to generate verbal reasoning traces and actions pertaining to a specific task in an interleaved manner while also allowing for interactions with specific environments 
that may be subsequently incorporated into reasoning. 
We also include systems such as BINDER~\citep{cheng2023binding} and ChatDB~\citep{hu2023chatdb} that interact with a Python interpreter for code execution or an SQL engine for  retrieval. 



\subsubsection*{Flexible Verbal Reasoning}

An alternate line of work explores orchestration that incorporates more flexibility into verbal reasoning, possibly by mimicking formal reasoning. 
Several state-of-the-art approaches use a \textit{branching structure} in decomposed prompts to guide the FM. 
The ADaPT system \citep{prasad2023adapt} allows in addition for task decomposition (or planning) to tackle more complex tasks.
Such a branching structure can appear in different forms, such as simple decomposition~\citep{radhakrishnan2023question}, 
tree-structures with search~\citep{yao2023tree}, 
or full graphs~\citep{besta2023graph}.

\subsubsection*{Formal Reasoning}

An upcoming class of approaches uses an FM to support a symbolic reasoner to solve a set of tasks.
Some examples include LM cascades~\citep{dohan2022language} and PLoT~\citep{wong2023word}, which both integrate FMs into a probabilistic programming language paradigm. These frameworks are described in more detail in a later section. 

%

%



We anticipate that all these research avenues will flourish in the future, with a growing interest in autonomous \texttt{Orchestrators} of various types, including work that leverages other kinds of formal reasoning (such as logical or decision-theoretic) for orchestration. 
Since \texttt{Orchestrators} play a special role in FM-guided AI systems, we rely on them heavily for the categorization in the next section.

\begin{table*}[t]
\centering
\scriptsize
\begin{tabular}{clcccccc}
\toprule
& Paper & FM Updater & Prompt Assistant & Assessor & Knowledge Curator & Orchestrator\\
\midrule
\multirow{5}{*}{\rotatebox[origin=c]{90}{\tiny  Update FM}}
& RLHF \citep{stiennon}  & Fine-tuning  & Instruction-based  & 
Binary Reward (Human)
& - & -  \\ 
& RLAIF \citep{bai2022constitutional}  & Fine-tuning  & In-context learning  & 
Harmfulness (Human)
& - & -  \\ 
& KG Infusion \citep{agarwal2021kelm}  & Modified Pre-training  & -  & -& Knowledge graphs & -  \\ 
& K-ADAPTER \citep{wang2021k} &Augmented Models  & -  & - & Wikipedia & -  \\ 
& MixDA \citep{diao2023mixtureofdomainadapters}  & Augmented Models & -  & - & Domain knowledge & -  \\ 
%
\midrule
\multirow{6}{*}{\rotatebox[origin=c]{90}{\tiny Access Tools}}
& RAG \citep{lewis_RAG}  & Controlled Generation  & -  & - & Wikipedia & -  \\ 
& IRCoT \citep{khot2022decomposed}  & -  & In-context Learning & - & Wikipedia/Corpus & -  \\ 
& ReAct \citep{yao2023react}  & Optional Fine-tuning  & In-context Learning  & Factuality (Verbal) & Web documents & API Calls  \\ 
& LLM Augmenter \citep{peng2023check}  & Controlled Generation  & Instruction-based  & Factuality (Human) & Web documents & API Calls  \\ 
& ChemCrow\citep{bran2023chemcrow}  & 
-
& In-context Learning  & Factuality (Human) & Chemistry expert knowledge  & API Calls  \\
& BINDER \citep{cheng2023binding}  & -  & In-context Learning   & Execute or fail (Formal) & 
-
& API Calls  \\ 
\midrule
\multirow{5}{*}{\rotatebox[origin=c]{90}{\tiny Dyn. Prompts}}
& ToT \citep{yao2023tree}  & -  & Decomposed Prompts  & Progress (Verbal) & - & Comb. search  \\ 
& GoT \citep{besta2023graph}  & -  & Decomposed Prompts  & Progress (Verbal) & - & Comb. search  \\ 
& ADaPT \citep{prasad2023adapt}  & -  & Decomposed Prompts  & Execute or fail (Verbal) & - & Comb. search  \\ 
& RAP \citep{hao2023reasoning} & -  & Decomposed Prompts  & Progress (Verbal) & - & Comb. search  \\
& Reflexion \citep{shinn2023reflexion}  & -  & Decomposed Prompts  & Progress (Verbal) & - & RL  \\ 
\midrule
\multirow{5}{*}{\rotatebox[origin=c]{90}{\tiny Ext. Reasoner}}
& LM Cascade \citep{dohan2022language}  & -  & In-context Learning  & Factuality (Formal) & 
-
& Prob. reasoning  \\ 
& CONCORD \citep{mitchell2022enhancing}  & -  & Instruction-based  & Consistency (Formal) & 
Other LMs
& 
-
\\ 
& PLoT \citep{wong2023word}  & -  & Instruction-based  & Factuality (Formal) & 
-
&Prob. reasoning  \\ 
& LARK \citep{choudhary2023complex}  & -  & Decomposed prompts  & - & Knowledge graphs & KG reasoning  \\ 
& CoK \citep{li2023chain}  & -  & Instruction-based  & Factuality (Formal) & External knowledge & Query languages  \\ 
\bottomrule
\end{tabular}
\caption{Illustrative frameworks for four categories of agent interaction protocols are shown in the rows. Values in the columns for five categories of sherpas are chosen as follows. For FM Updater and Prompt Assistant, categories are shown. For Assessor, the attribute being assessed is shown, along with the source of assessment (Human refers to human feedback, Verbal refers to direct assessment from an FM, and Formal refers to the use of formal reasoning). For Knowledge Curator, the source of the knowledge is shown. For Orchestrator, the main role/approach taken is shown.
}
\label{tab:your_label}
\end{table*}

\section{Agent Interaction Protocols}
\label{sec:Agent Interaction Protocols}

In this section, we categorize existing frameworks that guide FMs
based on \emph{interaction protocols} that we classify into four categories:
\textit{Update FM with External Knowledge}, 
\textit{Access Tools for Information Retrieval}, 
\textit{Explore Dynamic Prompts}, 
and
\textit{Integrate External Reasoners}.
These categories highlight common scenarios 
that span most current approaches. 

\subsection{Update FM with External Knowledge}
The simplest interaction protocol can be seen 
in frameworks for updating foundation models
with external knowledge from \texttt{Knowledge Curator} or 
feedback from users.
RLHF~\citep{stiennon} uses
human \texttt{Assessors} for the output of FMs, incorporating the feedback as a reward model for fine-tuning.
RLAIF~\citep{bai2022constitutional} extends RLHF 
by revising harmful responses from red teaming prompts
through a nuanced self-assessment of harmlessness 
based on user-provided constitutional principles
with \texttt{Prompt Assistant} that utilizes in-context learning.
Since fine-tuning a large FM is costly,
more recent work proposes augmenting smaller external models
without modifying the underlying FM~\citep{liu2024tuning,wang2021k,diao2023mixtureofdomainadapters}.
In this category, we observe that  \texttt{Orchestrator}
plays no significant role since interaction 
is a simple pre-defined workflow from 
\texttt{Knowledge Curator} or \texttt{Assessor}
to \texttt{FM Updater}.


\subsection{Access Tools for Information Retrieval}
%
For open-domain questions, 
FMs often show poor performance since it may be not possible to generate the desired answer solely from the input prompt.
Many practical frameworks enable retrieving relevant information from web documents or other application specific data through
pre-defined API calls~\citep{lewis_RAG,langchain}.
The overall agent interaction can be programmed in advance 
and \texttt{Prompt Assistant} equipped with static prompt templates can keep interacting with
\texttt{Knowledge Curator} and store retrieved information in external memory
to generate the final answer. 
The API calls can also be controlled by checking the factuality of 
the results, and \texttt{Orchestrator} can repeat the interaction
based on pre-defined programs.
\cite{peng2023check} propose a control flow similar to a reinforcement learning
agent that generates a feedback reward from a model-based \texttt{Assessor}
that evaluates factuality of the outcome.
\cite{bran2023chemcrow} explore organic synthesis, drug discovery, and material design by combining domain expertise with FMs.
User-specified scientific tasks are tackled by providing a sequence of prompts in a simple reasoning-based format to GPT-4, where one of 17 tools can be called upon; these could be as basic as web search, but also include \texttt{Assessors} for checking whether a molecule has been patented.

\subsection{Explore Dynamic Prompts}
%




In many few shot in-context learning prompting approaches,
an FM generates intermediate thoughts that lead to the answer following static examples provided in the input.
Searching for better combinations of intermediate thoughts,
\texttt{Orchestrator} coordinates \texttt{Prompt Assistant/Sequencer} to decompose the prompts and combine thoughts as a new input prompt 
for querying the FM, essentially implementing a combinatorial search 
over the space of all possible combinations of generated thoughts ~\citep{yao2023tree,besta2023graph}.
In addition to composing thoughts,
\cite{hao2023reasoning} and 
\cite{prasad2023adapt}
show how to define prompt templates for generating 
the state information in planning problems given a problem description
with the initial state, goal states, and set of actions.

Compared with the frameworks that access tools with a pre-defined interaction
programmed as a chain of static API calls,
the frameworks in this category exhibit more dynamic and complex interaction behavior. 
Many of the frameworks in this category
have several prompt templates that are associated with 
a component of search procedure 
such as node generation, node evaluation, or node selection.
A template of prompts is grounded to a fully instantiated prompt during search.
A particularly popular interaction protocol is that
\texttt{Prompt Assistant} generates successor states in the state space,
\texttt{Assessor} evaluates the quality of the state generated by the FM,
and
\texttt{Orchestrator} controls a search strategy such as 
Monte carlo tree search, local search, or a reinforcement learning control loop.
%

\subsection{Integrate External Reasoners}
%






All the previous categories primarily rely on the reasoning capability of an FM, aiming to address tasks by 
infusing additional domain knowledge, accessing external tools, or composing prompts to explore a suitable search space.
The last category follows an alternate approach that integrates external reasoning engines 
into \texttt{Assessor} or 
performs symbolic inference to extend \texttt{Knowledge Curator}
going beyond information retrieval.
For instance, 
\cite{dohan2022language} proposes language model cascades,
a probabilistic programming language paradigm 
that formulates natural language question-answering tasks 
as a graphical model over string-valued random variables associated 
with prompt inputs and FM outputs, 
thereby unifying various prompt-based reasoning approaches.
\cite{mitchell2022enhancing} also 
formulates a consistency \texttt{Assessor} as probabilistic inference.
PLoT \citep{wong2023word} integrates verbal reasoning by an FM
with symbolic reasoning by translating the natural language input as 
code using \texttt{Prompt Assistant} and solving an inference task using a probabilistic programming framework -- Church~\citep{goodman2008}.
There are frameworks 
that integrate reasoning systems other than probabilistic programming,
such as programming languages~\citep{cheng2023binding}
or knowledge graph reasoning~\citep{choudhary2023complex}.

\section{Discussion and Future Directions}

We discuss a vision where FM guiding sherpas 
have more autonomy, and highlight various topic areas for future research.



\subsection{Agent Behaviors for Intelligent FM Sherpas}

Much has been written about the properties of agents and their categorization
, as well as behaviors they need to exhibit to be deemed autonomous, smart, or intelligent. 
A typical definition for an intelligent agent is: ``a computer system situated in some environment and that is capable of flexible autonomous action in this environment in order to meet its design objectives''~\citep{jennings_book}. Considering a weak notion of agency~\citep{wooldridge_jennings_1995}, we envision agents exhibiting 
\emph{autonomy} (operating without the direct intervention of humans or other agents),
\emph{interactivity} (interacting with other agents including humans), \emph{reactivity} (responding to perceived changes in the environment), and \emph{pro-activity} (taking the initiative based on their motivations).
We also regard \emph{collaboration} (working together towards common goals) as a suitable endeavor.

While stronger notions of agency may be desirable for FM sherpas in the future, we believe that even weak agent behavior is beneficial for practical systems.
For instance, autonomy is a useful feature for an \texttt{Assessor}, enabling it to operate continuously without direct human intervention. 
An \texttt{Orchestrator} must interact with various other agents, including humans, since practical problems may need planning adjustments based on clarifications or requested feedback. 
Reactivity may also be key; consider an \texttt{Assessor} that needs to leverage changes in users or user personas. 
Another example is a pro-active \texttt{Knowledge Curator} that actively seeks out and reports appropriate knowledge in an enterprise, such as the possibility of regulatory changes. 

\subsection{Future Directions}

Although there have been significant recent advances around guiding FMs, numerous open problems and opportunities remain. Our proposed agent interaction categories can help inform future  protocols with more complex and nuanced interactions, including through sherpas that are equipped with stronger agency. Some potential directions follow:

\begin{itemize}[noitemsep,nolistsep,leftmargin=*]
\item \textit{Multi-objective and Joint Optimization Paradigms.}
Many interaction protocols are aimed at improving performance measures such as accuracy, but it may be desirable to optimize over a broader suite of objectives, such as run-time costs, level of harmfulness, or 
social bias, using quantified measures from \texttt{Assessors}. Also, it may be prudent to jointly optimize over multiple sherpas; for example, if human feedback is limited, there is a question around which sherpa that feedback may assist most meaningfully.
\item \textit{Broader Role for Automated Assessors.}
The interaction protocols for accessing tools or exploring dynamic prompts heavily rely on human feedback or verbal reasoning by FMs. Formal  \texttt{Assessors} could save the cost of collecting user feedback 
or replace less credible verbal reasoning  \texttt{Assessors}.
Furthermore, \texttt{FM Updaters} could directly query formal assessors and go beyond information retrieval to control generation of more trustworthy outputs during runtime. 
Similarly, \texttt{Prompt Assistants} could  dynamically generate optimized prompts by incorporating feedback from \texttt{Assessors} in an automated manner.

\item \textit{Uncertainty Quantification Assessors.}
While exploring the search space of dynamic prompts, there is potential in using
uncertainty quantification  \texttt{Assessors}
for informing search, so as to evaluate the predicted outcome or even prune the search space without wasting valuable resources by accessing FMs. 
\item \textit{Formal Reasoning Orchestration.}
There is scope for integrating formal reasoners directly
into dynamic \texttt{Prompt Assistants} 
to provide more accurate and diverse feedback to FMs.
\item \textit{Novel Modes of Knowledge Enablement.} Agents should be able to infuse knowledge from text, tabular data and graphs more seamlessly. 
Effective use of knowledge through \texttt{FM Updaters} and \texttt{Knowledge Curators} is still currently under explored, in our view. 
\item \textit{Humans in the Loop.}
The current literature merely scratches the surface of human involvement, such as for \texttt{Knowledge Curators} or \texttt{Assessors}. There is room for more sophisticated preference elicitation schemes from users that go beyond the `thumbs up/down' feedback approach of RLHF. 
Interaction of FM sherpas with humans through active learning will be a fruitful pursuit.
\item \textit{Benchmarks.}
There is a need for newer benchmarks that emphasize the breadth of a system's abilities as opposed to task-specific performance~\citep{ge2023openagi,yu2023kola}, 
which will undoubtedly result in significant innovations in sherpas and highlight their power.
\end{itemize}

\section{Conclusions}

%
%
In this paper,
we surveyed representative approaches for guiding FMs with knowledge and reasoning
%
using a conceptual framework
that elucidates the role of guiding agents. 
We categorized interaction protocols  into four major groups,
emphasizing the importance and   opportunities around
collaboration of agents with different roles.
%
%
Systems that can smartly and jointly leverage such guiding agents  would go a long way towards ensuring long-term usability and trustworthiness in real-world applications.

\clearpage

\bibliographystyle{named}



\footnotesize{

\bibliography{sherpas_survey}

\begin{thebibliography}{}

\bibitem[\protect\citeauthoryear{Agarwal \bgroup \em et al.\egroup
  }{2021}]{agarwal2021kelm}
Oshin Agarwal, Heming Ge, Siamak Shakeri, et~al.
\newblock Knowledge graph based synthetic corpus generation for
  knowledge-enhanced language model pre-training.
\newblock In {\em Proceedings of the Conference of the North American Chapter
  of the Association for Computational Linguistics: Human Language
  Technologies}, 2021.

\bibitem[\protect\citeauthoryear{Amodei \bgroup \em et al.\egroup
  }{2016}]{amodei2016concrete}
Dario Amodei, Chris Olah, Jacob Steinhardt, et~al.
\newblock Concrete problems in {AI} safety.
\newblock {\em arXiv:1606.06565}, 2016.

\bibitem[\protect\citeauthoryear{Arsenyan and
  Shahnazaryan}{2023}]{arsenyan2023large}
Vahan Arsenyan and Davit Shahnazaryan.
\newblock Large language models for biomedical causal graph construction.
\newblock {\em arXiv:2301.12473}, 2023.

\bibitem[\protect\citeauthoryear{Bai \bgroup \em et al.\egroup
  }{2022}]{bai2022constitutional}
Yuntao Bai, Saurav Kadavath, Sandipan Kundu, et~al.
\newblock Constitutional {AI}: Harmlessness from {AI} feedback.
\newblock {\em arXiv:2212.08073}, 2022.

\bibitem[\protect\citeauthoryear{Besta \bgroup \em et al.\egroup
  }{2023}]{besta2023graph}
Maciej Besta, Nils Blach, Ales Kubicek, et~al.
\newblock Graph of thoughts: Solving elaborate problems with large language
  models.
\newblock {\em arXiv:2308.09687}, 2023.

\bibitem[\protect\citeauthoryear{Bommasani \bgroup \em et al.\egroup
  }{2021}]{bommasani}
Rishi Bommasani, Drew~A. Hudson, Ehsan Adeli, et~al.
\newblock On the opportunities and risks of foundation models.
\newblock {\em arXiv:2108.07258}, 2021.

\bibitem[\protect\citeauthoryear{Bosselut \bgroup \em et al.\egroup
  }{2019}]{bosselut-etal-2019-comet}
Antoine Bosselut, Hannah Rashkin, Maarten Sap, et~al.
\newblock {COMET}: Commonsense transformers for automatic knowledge graph
  construction.
\newblock In {\em Proceedings of the Annual Meeting of the Association for
  Computational Linguistics}, 2019.

\bibitem[\protect\citeauthoryear{Bran \bgroup \em et al.\egroup
  }{2023}]{bran2023chemcrow}
Andres~M Bran, Sam Cox, Andrew~D White, and Philippe Schwaller.
\newblock Chemcrow: Augmenting large-language models with chemistry tools.
\newblock {\em arXiv:2304.05376}, 2023.

\bibitem[\protect\citeauthoryear{Chase}{2022}]{langchain}
Harrison Chase.
\newblock Langchain.
\newblock \url{https://github.com/langchain-ai/langchain}, October 2022.

\bibitem[\protect\citeauthoryear{Cheng \bgroup \em et al.\egroup
  }{2023}]{cheng2023binding}
Zhoujun Cheng, Tianbao Xie, Peng Shi, et~al.
\newblock Binding language models in symbolic languages.
\newblock In {\em International Conference on Learning Representations}, 2023.

\bibitem[\protect\citeauthoryear{Choudhary and
  Reddy}{2023}]{choudhary2023complex}
Nurendra Choudhary and Chandan~K Reddy.
\newblock Complex logical reasoning over knowledge graphs using large language
  models.
\newblock {\em arXiv:2305.01157}, 2023.

\bibitem[\protect\citeauthoryear{Cobbe \bgroup \em et al.\egroup
  }{2021}]{cobbe2021training}
Karl Cobbe, Vineet Kosaraju, Mohammad Bavarian, et~al.
\newblock Training verifiers to solve math word problems.
\newblock {\em arXiv:2110.14168}, 2021.

\bibitem[\protect\citeauthoryear{Creswell \bgroup \em et al.\egroup
  }{2022}]{creswell2022selection}
Antonia Creswell, Murray Shanahan, and Irina Higgins.
\newblock Selection-inference: Exploiting large language models for
  interpretable logical reasoning.
\newblock {\em arXiv:2205.09712}, 2022.

\bibitem[\protect\citeauthoryear{De~Cao \bgroup \em et al.\egroup
  }{2021}]{de2021editing}
Nicola De~Cao, Wilker Aziz, and Ivan Titov.
\newblock Editing factual knowledge in language models.
\newblock In {\em Empirical Methods in Natural Language Processing}, 2021.

\bibitem[\protect\citeauthoryear{Diao \bgroup \em et al.\egroup
  }{2023}]{diao2023mixtureofdomainadapters}
Shizhe Diao, Tianyang Xu, Ruijia Xu, et~al.
\newblock Mixture-of-domain-adapters: Decoupling and injecting domain knowledge
  to pre-trained language models memories.
\newblock {\em arXiv:2306.05406}, 2023.

\bibitem[\protect\citeauthoryear{Dohan \bgroup \em et al.\egroup
  }{2022}]{dohan2022language}
David Dohan, Winnie Xu, Aitor Lewkowycz, et~al.
\newblock Language model cascades.
\newblock In {\em Workshop Beyond Bayes: Paths Toward Universal Reasoning
  Systems}, 2022.

\bibitem[\protect\citeauthoryear{Dong \bgroup \em et al.\egroup
  }{2023}]{dong2023survey}
Qingxiu Dong, Lei Li, Damai Dai, et~al.
\newblock A survey on in-context learning.
\newblock {\em arXiv:2301.00234}, 2023.

\bibitem[\protect\citeauthoryear{Du \bgroup \em et al.\egroup
  }{2023}]{du2023improving}
Yilun Du, Shuang Li, Antonio Torralba, et~al.
\newblock Improving factuality and reasoning in language models through
  multiagent debate.
\newblock {\em arXiv:2305.14325}, 2023.

\bibitem[\protect\citeauthoryear{Elazar \bgroup \em et al.\egroup
  }{2021}]{elazar2021}
Yanai Elazar, Nora Kassner, Shauli Ravfogel, et~al.
\newblock Measuring and improving consistency in pretrained language models.
\newblock {\em Transactions of the Association for Computational Linguistics},
  2021.

\bibitem[\protect\citeauthoryear{Ge \bgroup \em et al.\egroup
  }{2023}]{ge2023openagi}
Yingqiang Ge, Wenyue Hua, Kai Mei, et~al.
\newblock {OpenAGI}: When {LLM} meets domain experts.
\newblock {\em arXiv:2304.04370}, 2023.

\bibitem[\protect\citeauthoryear{Goodman \bgroup \em et al.\egroup
  }{2008}]{goodman2008}
Noah~D. Goodman, Vikash~K. Mansinghka, Daniel Roy, et~al.
\newblock Church: a language for generative models.
\newblock In {\em Proceedings of the Conference on Uncertainty in Artificial
  Intelligence}, 2008.

\bibitem[\protect\citeauthoryear{Han \bgroup \em et al.\egroup
  }{2021}]{han2021pretrained}
Xu~Han, Zhengyan Zhang, Ning Ding, et~al.
\newblock Pre-trained models: Past, present and future.
\newblock {\em AI Open}, 2021.

\bibitem[\protect\citeauthoryear{Hao \bgroup \em et al.\egroup
  }{2023a}]{hao2023reasoning}
Shibo Hao, Yi~Gu, Haodi Ma, et~al.
\newblock Reasoning with language model is planning with world model.
\newblock In {\em Proceedings of the Conference on Empirical Methods in Natural
  Language Processing}, 2023.

\bibitem[\protect\citeauthoryear{Hao \bgroup \em et al.\egroup
  }{2023b}]{hao2023bertnet}
Shibo Hao, Bowen Tan, Kaiwen Tang, et~al.
\newblock {BertNet}: Harvesting knowledge graphs with arbitrary relations from
  pretrained language models.
\newblock {\em arXiv:2206.14268}, 2023.

\bibitem[\protect\citeauthoryear{Hu \bgroup \em et al.\egroup
  }{2023}]{hu2023chatdb}
Chenxu Hu, Jie Fu, Chenzhuang Du, et~al.
\newblock Chatdb: Augmenting llms with databases as their symbolic memory.
\newblock {\em arXiv:2306.03901}, 2023.

\bibitem[\protect\citeauthoryear{Huang and Chang}{2023}]{huang2023}
Jie Huang and Kevin Chen-Chuan Chang.
\newblock Towards reasoning in large language models: A survey.
\newblock In {\em Findings of Association for Computational Linguistics}, 2023.

\bibitem[\protect\citeauthoryear{Huang \bgroup \em et al.\egroup
  }{2023}]{huang2023survey}
Lei Huang, Weijiang Yu, Weitao Ma, et~al.
\newblock A survey on hallucination in large language models: Principles,
  taxonomy, challenges, and open questions.
\newblock {\em arXiv:2311.05232}, 2023.

\bibitem[\protect\citeauthoryear{Jennings and Wooldridge}{1998}]{jennings_book}
Nicholas~R. Jennings and Michael~J. Wooldridge, editors.
\newblock {\em Agent Technology: Foundations, Applications, and Markets}.
\newblock 1998.

\bibitem[\protect\citeauthoryear{Khot \bgroup \em et al.\egroup
  }{2023}]{khot2022decomposed}
Tushar Khot, Harsh Trivedi, Matthew Finlayson, et~al.
\newblock Decomposed prompting: A modular approach for solving complex tasks.
\newblock In {\em International Conference on Learning Representations}, 2023.

\bibitem[\protect\citeauthoryear{Kojima \bgroup \em et al.\egroup
  }{2022}]{kojima;neurips;2022}
Takeshi Kojima, Shixiang~(Shane) Gu, Machel Reid, et~al.
\newblock Large language models are zero-shot reasoners.
\newblock In {\em Advances in Neural Information Processing Systems}, 2022.

\bibitem[\protect\citeauthoryear{Kuhn \bgroup \em et al.\egroup
  }{2022}]{kuhn2022semantic}
Lorenz Kuhn, Yarin Gal, and Sebastian Farquhar.
\newblock Semantic uncertainty: Linguistic invariances for uncertainty
  estimation in natural language generation.
\newblock In {\em International Conference on Learning Representations}, 2022.

\bibitem[\protect\citeauthoryear{Kıcıman \bgroup \em et al.\egroup
  }{2023}]{kıcıman2023causal}
Emre Kıcıman, Robert Ness, Amit Sharma, and Chenhao Tan.
\newblock Causal reasoning and large language models: Opening a new frontier
  for causality.
\newblock {\em arXiv:2305.00050}, 2023.

\bibitem[\protect\citeauthoryear{Lewis \bgroup \em et al.\egroup
  }{2020}]{lewis_RAG}
Patrick Lewis, Ethan Perez, Aleksandra Piktus, et~al.
\newblock Retrieval-augmented generation for knowledge-intensive {NLP} tasks.
\newblock In {\em Advances in Neural Information Processing Systems}, 2020.

\bibitem[\protect\citeauthoryear{Li \bgroup \em et al.\egroup
  }{2023a}]{li2023camel}
Guohao Li, Hasan Abed Al~Kader Hammoud, Hani Itani, et~al.
\newblock Camel: Communicative agents for" mind" exploration of large scale
  language model society.
\newblock {\em arXiv:2303.17760}, 2023.

\bibitem[\protect\citeauthoryear{Li \bgroup \em et al.\egroup
  }{2023b}]{li2023chain}
Xingxuan Li, Ruochen Zhao, Yew~Ken Chia, et~al.
\newblock Chain of knowledge: A framework for grounding large language models
  with structured knowledge bases.
\newblock {\em arXiv:2305.13269}, 2023.

\bibitem[\protect\citeauthoryear{Lin \bgroup \em et al.\egroup
  }{2022}]{lin2022teaching}
Stephanie Lin, Jacob Hilton, and Owain Evans.
\newblock Teaching models to express their uncertainty in words.
\newblock {\em Transactions on Machine Learning Research}, 2022.

\bibitem[\protect\citeauthoryear{Liu \bgroup \em et al.\egroup
  }{2021}]{liu2021pretrain}
Pengfei Liu, Weizhe Yuan, Jinlan Fu, et~al.
\newblock Pre-train, prompt, and predict: A systematic survey of prompting
  methods in natural language processing.
\newblock {\em arXiv:2107.13586}, 2021.

\bibitem[\protect\citeauthoryear{Liu \bgroup \em et al.\egroup
  }{2024}]{liu2024tuning}
Alisa Liu, Xiaochuang Han, Yizhong Wang, et~al.
\newblock Tuning language models by proxy.
\newblock {\em arXiv:2401.08565}, 2024.

\bibitem[\protect\citeauthoryear{Long \bgroup \em et al.\egroup
  }{2023}]{long2023large}
Stephanie Long, Tibor Schuster, and Alexandre Piché.
\newblock Can large language models build causal graphs?
\newblock {\em arXiv:2303.05279}, 2023.

\bibitem[\protect\citeauthoryear{Mialon \bgroup \em et al.\egroup
  }{2023}]{mialon2023augmented}
Grégoire Mialon, Roberto Dessì, Maria Lomeli, et~al.
\newblock Augmented language models: A survey.
\newblock {\em arXiv:1606.06565}, 2023.

\bibitem[\protect\citeauthoryear{Min \bgroup \em et al.\egroup
  }{2022a}]{min;acl;2022}
Sewon Min, Mike Lewis, Hannaneh Hajishirzi, and Luke Zettlemoyer.
\newblock Noisy channel language model prompting for few-shot text
  classification.
\newblock In {\em Proceedings of the Association for Computational
  Linguistics}, 2022.

\bibitem[\protect\citeauthoryear{Min \bgroup \em et al.\egroup
  }{2022b}]{min;emnlp;2022}
Sewon Min, Xinxi Lyu, Ari Holtzman, et~al.
\newblock Rethinking the role of demonstrations: What makes in-context learning
  work?
\newblock In {\em Proceedings of the Conference on Empirical Methods in Natural
  Language Processing}, 2022.

\bibitem[\protect\citeauthoryear{Mitchell \bgroup \em et al.\egroup
  }{2022a}]{mitchell2022memory}
Eric Mitchell, Charles Lin, Antoine Bosselut, et~al.
\newblock Memory-based model editing at scale.
\newblock In {\em International Conference on Machine Learning}, 2022.

\bibitem[\protect\citeauthoryear{Mitchell \bgroup \em et al.\egroup
  }{2022b}]{mitchell2022enhancing}
Eric Mitchell, Joseph Noh, Siyan Li, et~al.
\newblock Enhancing self-consistency and performance of pre-trained language
  models through natural language inference.
\newblock In {\em Empirical Methods in Natural Language Processing}, 2022.

\bibitem[\protect\citeauthoryear{Moiseev \bgroup \em et al.\egroup
  }{2022}]{moiseev2022skill}
Fedor Moiseev, Zhe Dong, Enrique Alfonseca, and Martin Jaggi.
\newblock {SKILL}: Structured knowledge infusion for large language models.
\newblock In {\em North American Chapter of the Association for Computational
  Linguistics: Human Language Technologies}, 2022.

\bibitem[\protect\citeauthoryear{Pan \bgroup \em et al.\egroup
  }{2023}]{pan2023unifying}
Shirui Pan, Linhao Luo, Yufei Wang, et~al.
\newblock Unifying large language models and knowledge graphs: A roadmap.
\newblock {\em arXiv:2306.08302}, 2023.

\bibitem[\protect\citeauthoryear{Peng \bgroup \em et al.\egroup
  }{2023}]{peng2023check}
Baolin Peng, Michel Galley, Pengcheng He, et~al.
\newblock Check your facts and try again: Improving large language models with
  external knowledge and automated feedback.
\newblock {\em arXiv:2302.12813}, 2023.

\bibitem[\protect\citeauthoryear{Petroni \bgroup \em et al.\egroup
  }{2019}]{petroni2019language}
Fabio Petroni, Tim Rockt{\"a}schel, Sebastian Riedel, et~al.
\newblock Language models as knowledge bases?
\newblock In {\em Empirical Methods in Natural Language Processing and the
  International Joint Conference on Natural Language Processing}, 2019.

\bibitem[\protect\citeauthoryear{Prasad \bgroup \em et al.\egroup
  }{2023}]{prasad2023adapt}
Archiki Prasad, Alexander Koller, Mareike Hartmann, et~al.
\newblock Adapt: As-needed decomposition and planning with language models.
\newblock {\em arXiv:2311.05772}, 2023.

\bibitem[\protect\citeauthoryear{Qiao \bgroup \em et al.\egroup
  }{2023}]{qiao2023reasoning}
Shuofei Qiao, Yixin Ou, Ningyu Zhang, et~al.
\newblock Reasoning with language model prompting: A survey.
\newblock {\em arXiv:2212.09597}, 2023.

\bibitem[\protect\citeauthoryear{Radhakrishnan \bgroup \em et al.\egroup
  }{2023}]{radhakrishnan2023question}
Ansh Radhakrishnan, Karina Nguyen, Anna Chen, et~al.
\newblock Question decomposition improves the faithfulness of model-generated
  reasoning.
\newblock {\em arXiv:2307.11768}, 2023.

\bibitem[\protect\citeauthoryear{Rajani \bgroup \em et al.\egroup
  }{2019}]{rajani-etal-2019-explain}
Nazneen~Fatema Rajani, Bryan McCann, Caiming Xiong, and Richard Socher.
\newblock Explain yourself! leveraging language models for commonsense
  reasoning.
\newblock In {\em Annual Meeting of the Association for Computational
  Linguistics}, 2019.

\bibitem[\protect\citeauthoryear{Rubin \bgroup \em et al.\egroup
  }{2022}]{rubin-etal-2022-learning}
Ohad Rubin, Jonathan Herzig, and Jonathan Berant.
\newblock Learning to retrieve prompts for in-context learning.
\newblock In {\em North American Chapter of the Association for Computational
  Linguistics: Human Language Technologies}, 2022.

\bibitem[\protect\citeauthoryear{Shi \bgroup \em et al.\egroup
  }{2023}]{shi2023language}
Xiaoming Shi, Siqiao Xue, Kangrui Wang, et~al.
\newblock Language models can improve event prediction by few-shot abductive
  reasoning.
\newblock {\em arXiv:2305.16646}, 2023.

\bibitem[\protect\citeauthoryear{Shinn \bgroup \em et al.\egroup
  }{2023}]{shinn2023reflexion}
Noah Shinn, Federico Cassano, Ashwin Gopinath, et~al.
\newblock Reflexion: Language agents with verbal reinforcement learning.
\newblock In {\em Advances in Neural InformationProcessing Systems}, 2023.

\bibitem[\protect\citeauthoryear{Stiennon \bgroup \em et al.\egroup
  }{2020}]{stiennon}
Nisan Stiennon, Long Ouyang, Jeff Wu, et~al.
\newblock Learning to summarize from human feedback.
\newblock In {\em Advances in Neural Information Processing Systems}, 2020.

\bibitem[\protect\citeauthoryear{Wang \bgroup \em et al.\egroup
  }{2021}]{wang2021k}
Ruize Wang, Duyu Tang, Nan Duan, et~al.
\newblock K-adapter: Infusing knowledge into pre-trained models with adapters.
\newblock In {\em Findings of the Association for Computational Linguistics},
  2021.

\bibitem[\protect\citeauthoryear{Wei \bgroup \em et al.\egroup
  }{2021}]{wei2021knowledge}
Xiaokai Wei, Shen Wang, Dejiao Zhang, et~al.
\newblock Knowledge enhanced pretrained language models: A compreshensive
  survey.
\newblock {\em arXiv:2110.08455}, 2021.

\bibitem[\protect\citeauthoryear{Wei \bgroup \em et al.\egroup
  }{2022}]{wei2023chainofthought}
Jason Wei, Xuezhi Wang, Dale Schuurmans, et~al.
\newblock Chain-of-thought prompting elicits reasoning in large language
  models.
\newblock In {\em Advances in Neural Information Processing Systems}, 2022.

\bibitem[\protect\citeauthoryear{West \bgroup \em et al.\egroup
  }{2022}]{west_distillation}
Peter West, Chandra Bhagavatula, Jack Hessel, et~al.
\newblock Symbolic knowledge distillation: From general language models to
  commonsense models.
\newblock In {\em North American Chapter of the Association for Computational
  Linguistics: Human Language Technologies}, 2022.

\bibitem[\protect\citeauthoryear{Wong \bgroup \em et al.\egroup
  }{2023}]{wong2023word}
Lionel Wong, Gabriel Grand, Alexander~K Lew, et~al.
\newblock From word models to world models: Translating from natural language
  to the probabilistic language of thought.
\newblock {\em arXiv:2306.12672}, 2023.

\bibitem[\protect\citeauthoryear{Wooldridge and
  Jennings}{1995}]{wooldridge_jennings_1995}
Michael Wooldridge and Nicholas~R. Jennings.
\newblock Intelligent agents: Theory and practice.
\newblock {\em The Knowledge Engineering Review}, 1995.

\bibitem[\protect\citeauthoryear{Wu \bgroup \em et al.\egroup
  }{2023}]{wu2023autogen}
Qingyun Wu, Gagan Bansal, Jieyu Zhang, et~al.
\newblock Autogen: Enabling next-gen llm applications via multi-agent
  conversation framework.
\newblock {\em arXiv:2308.08155}, 2023.

\bibitem[\protect\citeauthoryear{Yang \bgroup \em et al.\egroup
  }{2023}]{yang2023automatonbased}
Yunhao Yang, Jean-Raphaël Gaglione, Cyrus Neary, and Ufuk Topcu.
\newblock Automaton-based representations of task knowledge from generative
  language models.
\newblock {\em arXiv:2212.01944}, 2023.

\bibitem[\protect\citeauthoryear{Yao \bgroup \em et al.\egroup
  }{2023a}]{yao2023tree}
Shunyu Yao, Dian Yu, Jeffrey Zhao, et~al.
\newblock Tree of thoughts: Deliberate problem solving with large language
  models.
\newblock {\em arXiv:2305.10601}, 2023.

\bibitem[\protect\citeauthoryear{Yao \bgroup \em et al.\egroup
  }{2023b}]{yao2023react}
Shunyu Yao, Jeffrey Zhao, Dian Yu, et~al.
\newblock React: Synergizing reasoning and acting in language models.
\newblock In {\em International Conference on Learning Representations}, 2023.

\bibitem[\protect\citeauthoryear{Yu \bgroup \em et al.\egroup
  }{2023}]{yu2023kola}
Jifan Yu, Xiaozhi Wang, Shangqing Tu, et~al.
\newblock {KoLA}: Carefully benchmarking world knowledge of large language
  models.
\newblock {\em arXiv:2306.09296}, 2023.

\bibitem[\protect\citeauthoryear{Zelikman \bgroup \em et al.\egroup
  }{2022}]{zelikman2022star}
Eric Zelikman, Yuhuai Wu, Jesse Mu, and Noah Goodman.
\newblock Star: Bootstrapping reasoning with reasoning.
\newblock {\em Advances in Neural Information Processing Systems}, 2022.

\bibitem[\protect\citeauthoryear{Zhao \bgroup \em et al.\egroup
  }{2023a}]{zhao2023explainability}
Haiyan Zhao, Hanjie Chen, Fan Yang, et~al.
\newblock Explainability for large language models: A survey.
\newblock {\em ACM Transactions on Intelligent Systems and Technology}, 2023.

\bibitem[\protect\citeauthoryear{Zhao \bgroup \em et al.\egroup
  }{2023b}]{zhao2023gptbias}
Jiaxu Zhao, Meng Fang, Shirui Pan, et~al.
\newblock {GPTBIAS}: A comprehensive framework for evaluating bias in large
  language models.
\newblock {\em arXiv:2312.06315}, 2023.

\bibitem[\protect\citeauthoryear{Zhen \bgroup \em et al.\egroup
  }{2022}]{zhen2022}
Chaoqi Zhen, Yanlei Shang, Xiangyu Liu, et~al.
\newblock A survey on knowledge-enhanced pre-trained language models.
\newblock {\em arXiv:2212.13428}, 2022.

\bibitem[\protect\citeauthoryear{Zhong \bgroup \em et al.\egroup
  }{2023}]{zhong2023mquake}
Zexuan Zhong, Zhengxuan Wu, Christopher~D Manning, et~al.
\newblock {MQuAKE}: Assessing knowledge editing in language models via
  multi-hop questions.
\newblock {\em arXiv:2305.14795}, 2023.

\bibitem[\protect\citeauthoryear{Zhou \bgroup \em et al.\egroup
  }{2023}]{zhou2023least}
Denny Zhou, Nathanael Sch{\"a}rli, Le~Hou, et~al.
\newblock Least-to-most prompting enables complex reasoning in large language
  models.
\newblock In {\em International Conference on Learning Representations}, 2023.

\end{thebibliography}

}

\end{document}